\documentclass[letterpaper, 10 pt, conference]{icra_template/ieeeconf}  
\makeatletter
\@ifundefined{NAT@parse}{}{\let\NAT@parse\@undefined}
\makeatother

\let\labelindent\relax

\usepackage{graphicx}
\usepackage[table]{xcolor}
\usepackage{amsfonts}
\usepackage{amsmath}
\usepackage{comment}
\usepackage{caption} 
\usepackage{microtype} 
\usepackage{hyperref} 
\usepackage{float} 

\usepackage{booktabs}

\makeatletter
\renewcommand{\section}{%
  \@startsection{section}{1}{\z@}%
                {-1.6ex \@plus -0.4ex \@minus -0.2ex}%
                { 0.9ex \@plus  0.2ex \@minus  0.1ex}%
                {\large\bf\raggedright}%
}
\renewcommand{\subsection}{%
  \@startsection{subsection}{2}{\z@}%
                {-1.0ex \@plus -0.3ex \@minus -0.2ex}%
                { 0.35ex \@plus  0.15ex}%
                {\normalsize\bf\raggedright}%
}
\renewcommand{\subsubsection}{%
  \@startsection{subsubsection}{3}{\z@}%
                {-0.85ex \@plus -0.25ex \@minus -0.2ex}%
                { 0.25ex \@plus  0.1ex}%
                {\normalsize\bf\raggedright}%
}
\makeatother

\providecommand{\keywords}[1]{\par\noindent\textbf{Keywords: }#1\par}

\usepackage{cuted}
\makeatletter
\def\@IEEEdynamictitlevspace{\vspace{0pt}}
\makeatother

\newcommand{\papername}{WeaveRL}%
\title{\papername: {Weaving Reconstruction into Scene-Aware Fabrics for Perceptive Reinforcement Learning}}

\author{
  \textbf{Remo Steiner}\textsuperscript{*}, \textbf{Vikram Ramasamy}\textsuperscript{*},
  \textbf{David Tingdahl}, \textbf{Sam Mady}, \textbf{Karl Van Wyk}, \\
  \textbf{Nathan Ratliff}, \textbf{David Recasens}, \textbf{Soha Pouya}, \textbf{Tuur Stuyck}, \textbf{Alexander Millane} \\
  NVIDIA Corporation. Zurich, Seattle and Santa Clara. \\
  *Equal Contribution \\
}

\usepackage{nicefrac}
\usepackage{pifont}
\usepackage{upgreek}
\usepackage[normalem]{ulem}
\usepackage{enumitem} 

\definecolor{MyDarkBlue}{rgb}{0,0.08,1}
\definecolor{MyLightBlue}{RGB}{55, 140, 231}
\definecolor{MyDarkGreen}{rgb}{0.02,0.6,0.02}
\definecolor{MyDarkRed}{rgb}{0.8,0.02,0.02}
\definecolor{MyDarkOrange}{rgb}{0.40,0.2,0.02}
\definecolor{MyLightOrange}{rgb}{0.90,0.5,0.02}
\definecolor{MyPurple}{RGB}{111,0,255}
\definecolor{MyLightPurple}{RGB}{126, 96, 191}
\definecolor{MyRed}{rgb}{1.0,0.0,0.0}
\definecolor{MyGold}{rgb}{0.75,0.6,0.12}
\definecolor{MyDarkgray}{rgb}{0.66, 0.66, 0.66}
\definecolor{MyLightGray}{rgb}{0.8, 0.8, 0.8}
\definecolor{MyWineRed}{rgb}{0.694,0.071, 0.149}
\definecolor{nicegreen}{rgb}{0.1, 0.6, 0.2}

\newcommand{\nvblox}{\textit{nvblox}}

\begin{document}
\bstctlcite{BSTcontrol}

\maketitle

\pagenumbering{arabic}
\pagestyle{plain}
\thispagestyle{plain}



\begingroup
\setlength{\abovecaptionskip}{3pt}
\setlength{\belowcaptionskip}{6pt}

\vspace{-18pt}
\begin{strip}
  \centering
  \begin{minipage}[t]{0.49\textwidth}
    \centering
    \includegraphics[width=\linewidth]{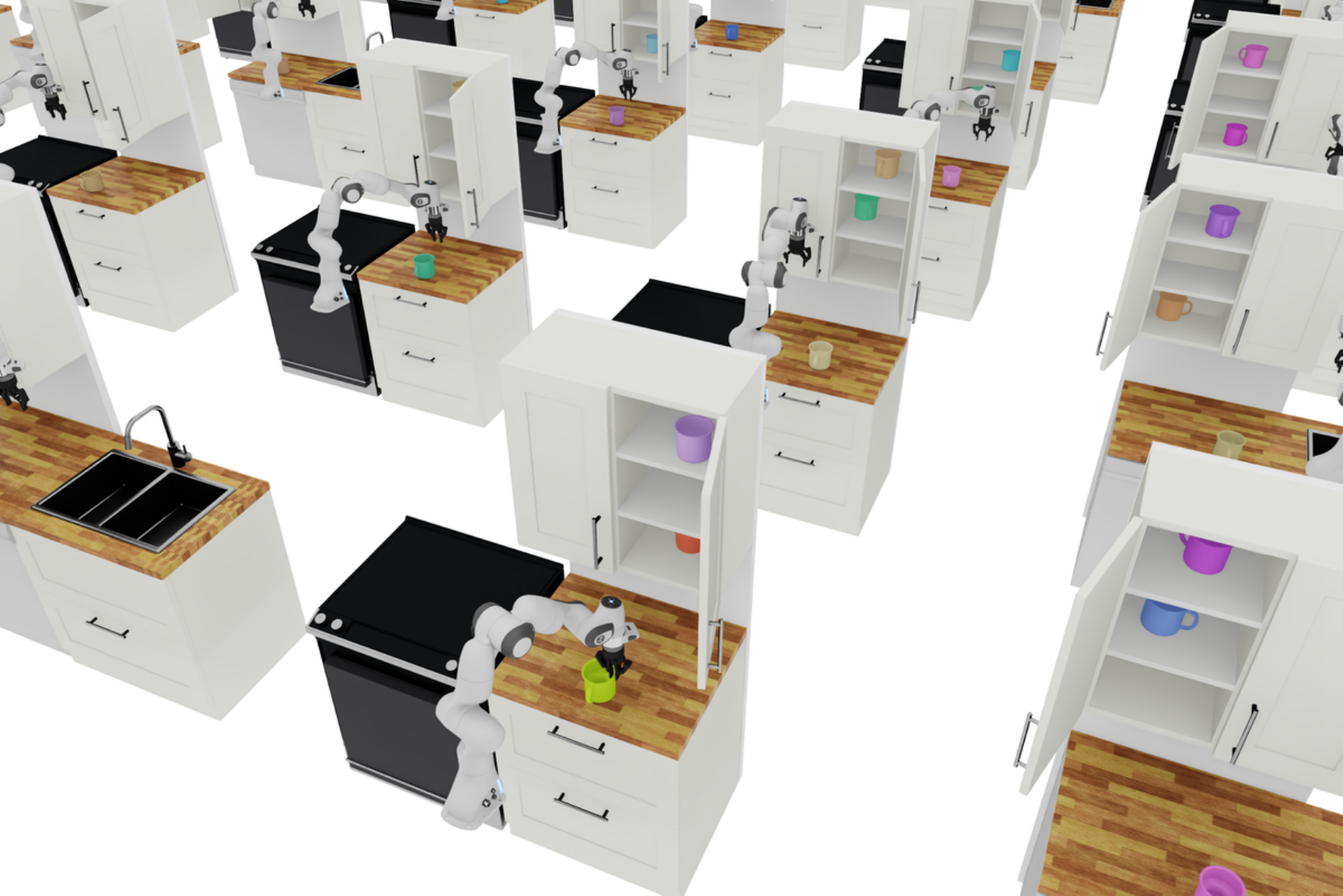}

  \end{minipage}\hfill%
  \begin{minipage}[t]{0.49\textwidth}
    \centering
    \includegraphics[width=\linewidth]{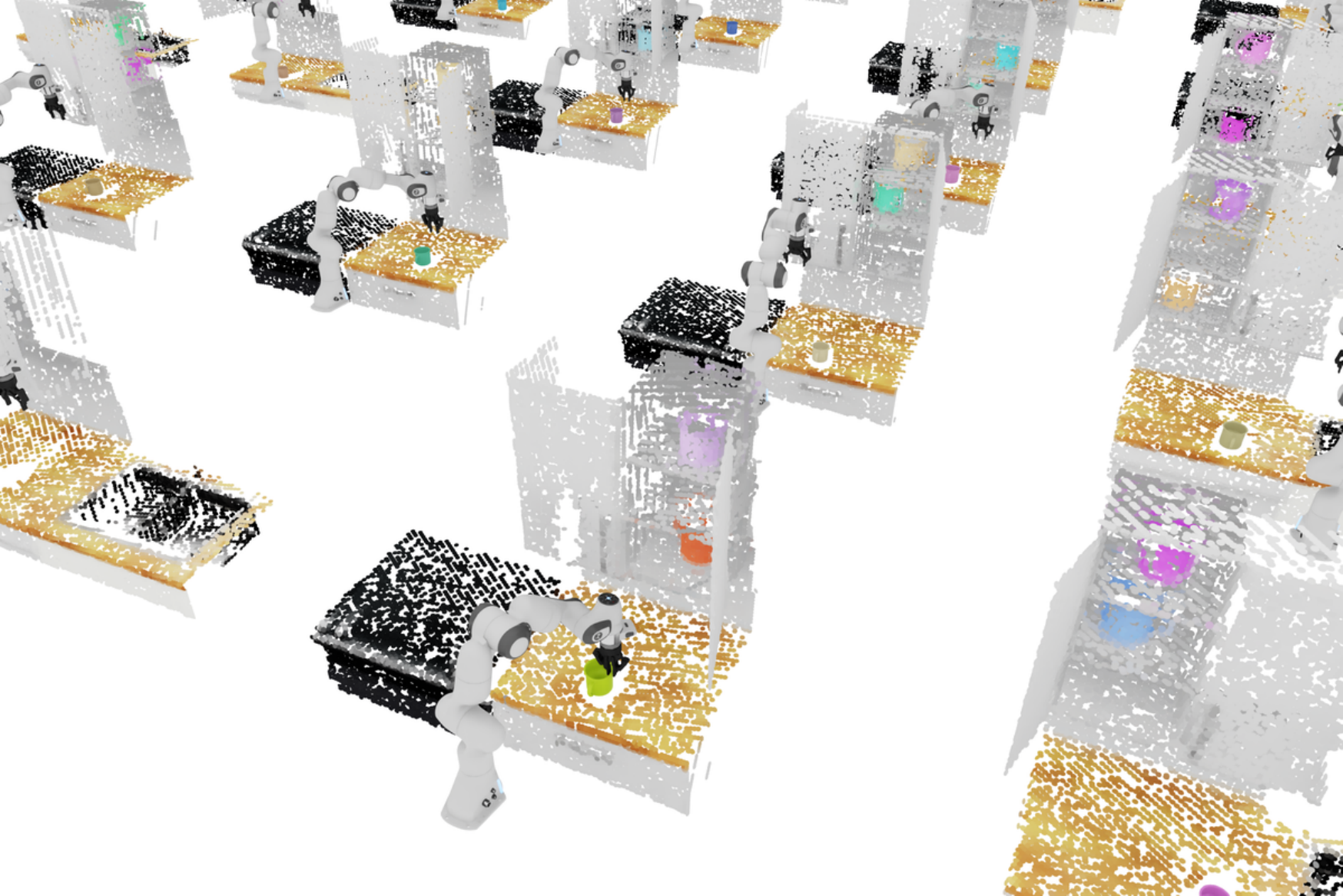}
  \end{minipage}

   \captionof{figure}{\textbf{Highly parallel scene reconstruction for reinforcement learning.} Rendering of RL kitchen environments for a task in which a robot places a mug in a wall-mounted cabinet (\emph{left}). Our tensorized surfel reconstruction system (\emph{right}) enables high-throughput mapping across thousands of parallel RL environments. Using scene-aware geometric fabrics, our method translates end-effector goals from the learned policy into collision-avoidant joint trajectories for collision-dense manipulation.}
  \label{fig:teaser}
\end{strip}

\vspace{6pt}

\endgroup


\begin{abstract}

Reinforcement learning allows robots to acquire complex skills, but producing policies for geometrically complex manipulation remains difficult. A promising approach is to learn on top of collision-avoidant controllers, such as geometric fabrics. However, these approaches have relied on static, hand-specified representations of the scene.
Integrating active, online 3D perception into massively parallel RL training has so far been inaccessible.
We introduce a GPU-accelerated method that reconstructs the scene as a collection of surfels across thousands of parallel simulation instances during active rollouts. This lets policies operate over sensor-derived, rather than hand-specified, geometry. On a suite of collision-dense manipulation tasks, our surfel fabrics enable policies to tackle geometrically complex scenes where primitive-based baselines fail, while maintaining sim-to-real transfer. Furthermore, policies learned with a scene-aware fabric are more robust to the introduction of novel geometry at test time, improving collision-free task completion under unseen obstacles from 35\% to 61\%. We release our reconstruction system, training code and test dataset to spur research in this direction\footnote{\url{https://weave-rl.github.io}}.

\end{abstract}

\section{Introduction}
\label{sec:introduction}

As robots are deployed into complex, unstructured environments like factories, homes, and hospitals, the ability to reason about scene geometry becomes critical. In recent years, Reinforcement Learning (RL) has produced state-of-the-art results for acquiring complex skills, revolutionizing robotic locomotion. RL is now expanding into complex manipulation tasks, which require vision-based sensing for policies to reason about the world they interact with. Building world representations that are sufficiently efficient for robot learning remains an open question.

The dominant paradigm in perceptive RL is to learn end-to-end visuomotor policies that map raw observations (e.g. pixels or pointclouds) directly to low-level control commands~\cite{levine2016end, kalashnikov2018qtopt, qin2023dexpoint}. While flexible, these architectures conflate visual perception, geometric reasoning, and motor control into a single model.
An alternative line of work decouples control from spatial awareness, pairing a higher-level learned policy with structured control frameworks such as geometric fabrics~\cite{van2022geometric} and explicit environment geometry. 
These approaches, however, face significant limitations.
They either represent the world as hand-engineered geometric primitives (e.g., bounding spheres or cuboids) that fail to capture complex, real-world scenes, or consume pre-computed, static environment maps that do not react to changes in the scene.
This presents a trade-off between these two paradigms: rely directly on raw sensor measurements and end-to-end learning, or use an explicit 3D map with limited expressivity and reactivity. In this work, we alleviate this trade-off.

We present a GPU-accelerated reconstruction pipeline designed for batched execution across parallel simulation instances.
By performing reconstruction during active RL rollouts, our method provides explicit, high-fidelity geometric representations of sensed environments that can be used by the policy for collision avoidance, without sacrificing the computational throughput necessary for modern RL.
We demonstrate the efficacy of our approach by extending the DextrAH framework~\cite{lum2024dextrah, singh2024dextrah} and replacing primitive-based world models with our real-time reconstructions.
We show that training time reconstruction is both possible with modest computational costs and that this approach generates policies that extend to complex tasks and are more robust to test time variations.
In summary, our key contributions are:

\begin{itemize}[leftmargin=*, noitemsep]
    \item \textbf{Parallelized, GPU-Accelerated 3D Reconstruction:} A
    tensorized surfel reconstruction pipeline, which enables high-throughput mapping across
    thousands of parallel environments during active RL rollouts.
    \item \textbf{Reconstruction-in-the-Loop RL System:} To our
    knowledge, the first RL framework where policies are trained,
    distilled, and deployed using a shared, real-time 3D
    reconstruction.
    \item \textbf{Scene-Aware Geometric Fabrics:} We extend the collision-avoidant behavior of geometric fabrics to cluttered scenes by replacing
    hand-engineered primitives with dynamic, sensor-derived geometry.
    \item \textbf{Enhanced Policy Robustness and Generalization:}
    Policies trained with our method are more resilient
    to test time variations than baseline methods.
\end{itemize}

\section{Related Work}

We present a unified framework that tightly couples a novel, batched 3D reconstruction pipeline with geometric fabrics to train, distill, and deploy manipulation policies over a live spatial representation. Below, we situate our approach within prior work on batchable reconstruction, structured control, and perceptive policies.

\paragraph{Batchable 3D Reconstruction for RL}
Constructing a world model from raw sensory inputs at the scale required for RL training remains a fundamental bottleneck, and existing representations address this challenge only partially. 
Pointcloud policies~\cite{drp2025} react to a 3D world representation, but only at the current instant, with no memory of what is occluded, leading to surprising limitations in state-of-the-art models~\cite{steiner2025mindmap}.
Mapping libraries have used surfels~\cite{whelan2015elasticfusion}, distance fields~\cite{millane2024nvblox, newcombe2011kinectfusion}, pointclouds~\cite{zhang2014loam}, and occupancy grids~\cite{hornung2013octomap} to accumulate geometry over time, but were built for deployment: a single robot in a single scene, outside any training loop.
By contrast, our system is batchable across the thousands of environments that RL runs in parallel.
This allows, for the first time, reconstructions to be built and queried online inside large-scale RL training.

\paragraph{Environment Representations for Structured Control}
In structured control, a learned policy steers an underlying controller that exposes a safe action space whose geometry encodes constraints (e.g., collisions), allowing the policy to focus on task completion rather than low-level motion generation and obstacle avoidance. The most developed controllers of this kind, such as RMPs, RMPflow, and geometric fabrics~\cite{ratliff2018rmp, ratliff2018rmpflow, van2022geometric, van2024geometric}, compose behaviors into stable, reactive motion. A critical limitation of these frameworks lies in how the environment's geometry is provided. Traditionally, this geometry must be explicitly hand-specified as geometric primitives. For example, DextrAH~\cite{lum2024dextrah, singh2024dextrah} and SafePBDS~\cite{wu2026safe} demonstrate the strength of this recipe for dexterous and certifiably safe grasping, yet they rely on a hand-crafted world.
Spahn et al.~\cite{spahn2025implicit} show that geometric fabrics can instead operate on online, sensor-derived scene representations including signed distance fields (SDFs) and raw pointclouds. However, their experiments use manually specified goal-attraction, rather than training an RL policy to provide task-level commands through the fabric. In related recent work, CSSDF-Net~\cite{cssdfnet} pre-trains a model to predict configuration-space collision distance and gradient from obstacle points for use in online optimization/MPC. In contrast, we train an RL policy to control a fabric over live, sensor-derived geometry that is batched across parallel environments during both training and deployment.

\paragraph{Reactive Neural Motion Policies}
Another approach is to learn policies end-to-end that
map sensor observations directly to motion commands, folding
perception, geometry, and control into a single model. Flow Motion
Policy~\cite{flowmotion} is an open-loop neural motion planner that samples
candidate paths with flow matching, collision checks them after generation,
and selects the first collision-free trajectory. However, a finite sample
set cannot guarantee a collision-free candidate. Deep Reactive
Policy~\cite{drp2025} requires pretraining an end-to-end pointcloud policy
on over 10 million expert trajectories, followed by distillation from a
geometric fabric teacher for static obstacle avoidance. In contrast, our
method retains the same geometric fabric throughout RL training and
deployment, and is trained from scratch without pre-training.

\section{Method}
\label{sec:method}

\begin{figure*}[t]
    \centering
    \includegraphics[width=\textwidth]{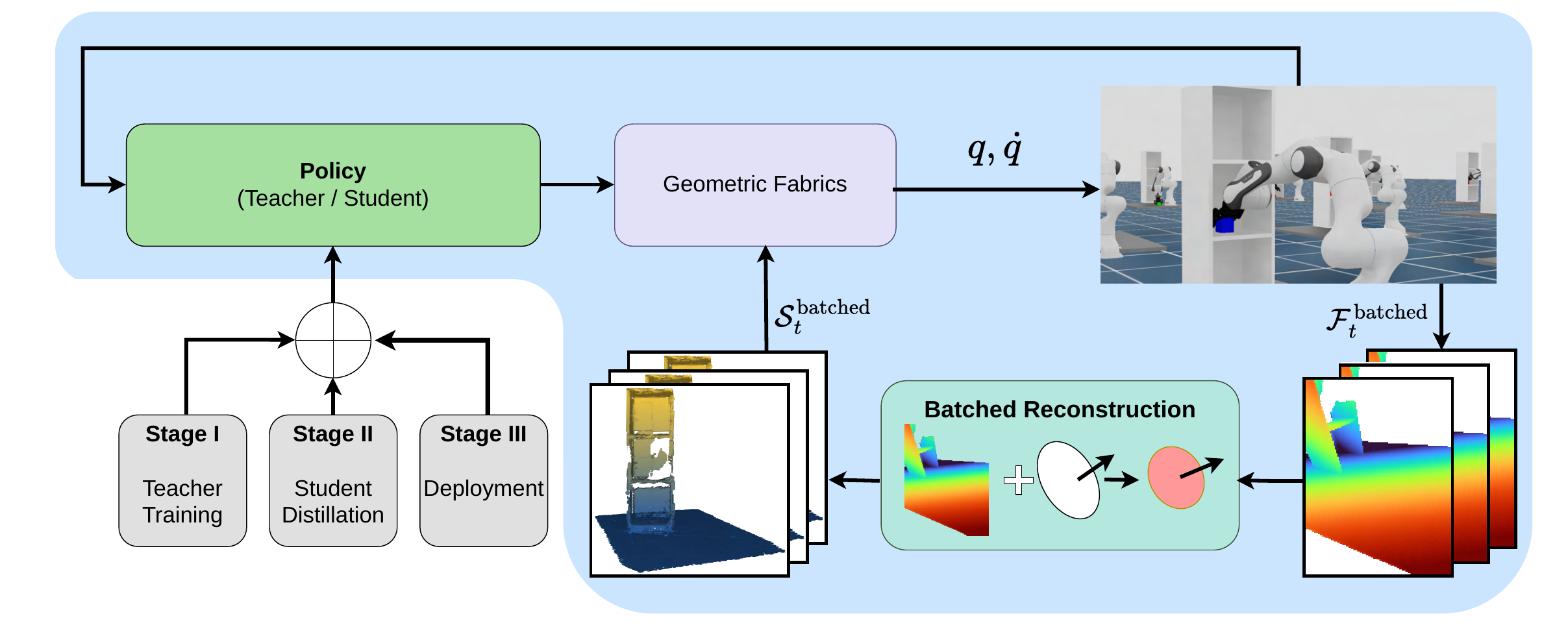}
    \caption{\textbf{WeaveRL Architecture.} At each control step, the policy receives a batch of images from the environment and computes targets for the geometric fabric, which emits joint commands $q,\dot{q}$. Batched depth frames $\mathcal{F}_t^{\mathrm{batched}}$ are fused into surfel maps $\mathcal{S}_t^{\mathrm{batched}}$, which the fabric queries on the next step for collision avoidance. Batched reconstruction across parallel environments enables the same perception--control loop during teacher RL training, student distillation, and deployment.}
    \vspace{-10pt}
    \label{fig:overview}
\end{figure*}

We present a GPU-accelerated, batched 3D vision pipeline for high-fidelity surfel reconstruction across parallel instances (Sec.~\ref{subsec:recon}), and integrate it with geometric fabrics by feeding the resulting live, sensor-derived geometry into the fabric at each control step (Sec.~\ref{subsec:manip}). Fig.~\ref{fig:overview} provides an overview.

\subsection{Parallelized 3D Reconstruction for Batched Environments}
\label{subsec:recon}

We build upon the surfel (surface element) representation, originally introduced as a zero-connectivity rendering primitive~\cite{pfister2000surfels} and later adapted for dense, point-based spatial fusion by systems such as ElasticFusion~\cite{whelan2015elasticfusion} and scalable CPU reconstruction pipelines~\cite{wang2019real}. At time $t$, the active spatial state of an environment is defined by a continuous map of oriented discs, $\mathcal{S}_{t}=\{\mathbf{s}_{i}=(\mathbf{p}_{i},\mathbf{n}_{i},r_{i})\}$, where $\mathbf{p}_{i}\in\mathbb{R}^{3}$ is the center, $\mathbf{n}_{i}\in\mathbb{S}^{2}$ is the unit normal, and $r_{i}$ is the radius. Our primary contribution is an architectural redesign that makes this process fast, batchable, and differentiable.

To support reconstruction during $N_b$ parallel RL rollouts simultaneously, we define a batched spatial state $\mathcal{S}_{t}^{\text{batched}}=\{\mathcal{S}_{t}^{(b)}\}_{b=1}^{N_b}$. Rather than maintaining dynamic, heterogeneous lists for each environment, which introduces memory fragmentation and branching, we encapsulate the entire batched map within fixed-shape tensors. Each geometric attribute is allocated as a tensor of shape $[N_b, C, F]$, where $C$ represents the preallocated surfel capacity and $F$ denotes the attribute dimensionality (e.g., $\mathbf{P} \in \mathbb{R}^{N_b \times C \times 3}$ for surfel centers). A corresponding boolean mask matrix $\mathbf{B} \in \{0,1\}^{N_b \times C}$ tracks occupancy of valid surfels within the block, keeping tensor shapes fixed across parallel maps so the full fusion pipeline can be compiled into fused GPU kernels. If a map outgrows its current capacity, we reallocate with a expansion factor to reduce the frequency of resizing and kernel compilation.

Given a batch of incoming RGB-D sensor observations $\mathcal{F}_{t}^{\text{batched}}=\{\mathcal{F}_{t}^{(b)}\}_{b=1}^{N_b}$, unprojection, data association, and state updates are executed via vectorized operations along the batch dimension. We maintain a compacted layout where valid surfels occupy indices $[0, N_{\mathrm{valid}})$, so newly observed geometry can be appended efficiently at $[N_{\mathrm{valid}},\, N_{\mathrm{valid}}+N_{\mathrm{new}})$ via pointer arithmetic. Furthermore, implementing the fusion and surface rendering directly in standard deep learning frameworks renders the pipeline fully differentiable, enabling potential end-to-end gradient flow from the spatial representation back to upstream perception modules (e.g., image encoders and sensor models).

\subsection{Collision-Dense Manipulation with Scene-Aware Fabrics}
\label{subsec:manip}

The ability to simultaneously maintain thousands of independent, high-fidelity spatial states across parallel environments unlocks novel capabilities. We build upon the geometric fabric formulation (refer to Appendix B in~\cite{adept2026} for a more complete description), a family of second-order reactive controllers that synthesize independent task behaviors into a stable, collision-aware dynamical system. We substitute the hand-crafted spatial priors of prior fabric-guided
policies with the live, sensor-derived batched spatial state described in Sec.~\ref{subsec:recon}.

Consider a highly articulated robotic manipulator tasked with producing collision-free, goal-driven motion within an unstructured, actively perceived environment. Let $q \in \mathcal{C}$ denote the robot's configuration in joint space, with $\dot{q}$ and $\ddot{q}$ representing its instantaneous velocity and acceleration, respectively. At each discrete control step $t$, the batched reconstruction pipeline provides an updated surfel map $\mathcal{S}_{t}=\{\mathbf{s}_{i}=(\mathbf{p}_{i},\mathbf{n}_{i},r_{i})\}$. To synthesize the obstacle repulsion task, the fabric requires a continuous Euclidean distance field $\Phi_{t}:\mathbb{R}^{3}\rightarrow\mathbb{R}$ and its corresponding spatial gradient $\nabla\Phi_{t}$. The robot's physical volume is conservatively approximated by a set of bounding spheres attached along its kinematic chain, with forward-kinematic centers $\mathbf{x}_{k}(q)$ and radii $\rho_{k}$. For a query point $\mathbf{x} \in \mathbb{R}^{3}$ and a specific surfel $s_{i}$, we define the orthogonal distance to the surfel's tangent plane as $h_{i}(\mathbf{x})=\mathbf{n}_{i}^{\top}(\mathbf{x}-\mathbf{p}_{i})$ and the in-plane radial offset as $\rho_{i}(\mathbf{x})=\lVert (\mathbf{x}-\mathbf{p}_{i})-h_{i}(\mathbf{x})\mathbf{n}_{i}\rVert$. The Euclidean distance $d_{i}(\mathbf{x})$ from the query point to the bounded surfel disc is given by:
\begin{equation}
  d_i(\mathbf{x}) =
  \begin{cases}
    |h_i(\mathbf{x})|, & \rho_i(\mathbf{x}) \le r_i,\\[2pt]
    \sqrt{h_i(\mathbf{x})^2 + \big(\rho_i(\mathbf{x}) - r_i\big)^2}, & \rho_i(\mathbf{x}) > r_i.
  \end{cases}
  \label{eq:disc_dist}
\end{equation}

The global distance field of the reconstructed environment is defined as the point-wise minimum across all valid surfels, $\Phi_{t}(\mathbf{x})=\min_{i}d_{i}(\mathbf{x})$. To determine the direction of retreat, we compute the spatial gradient, $\nabla\Phi_{t}(\mathbf{x})$, defined as the unit vector originating from the closest point on the nearest surfel pointing toward the query point. Because both $\Phi_{t}$ and $\nabla\Phi_{t}$ admit closed-form, differentiable solutions per surfel, the field can be highly vectorized and evaluated at every collision sphere center with limited computational overhead (see Sec~\ref{sec:timings-breakdown}).

The obstacle avoidance task is constructed independently for each collision sphere, driven directly by the geometric relationship to the nearest surfel. We evaluate the free distance $d_{k}=\Phi_{t}(\mathbf{x}_{k})-\rho_{k}$ and the retreat direction $\hat{\mathbf{u}}_{k}=\nabla\Phi_{t}(\mathbf{x}_{k})$. The fabric is subsequently supplied with a repulsive acceleration $\ddot{\mathbf{x}}_{k}^{rep}=\alpha(d_{k})\hat{\mathbf{u}}_{k}$, where the distance-based activation function $\alpha(d)$ equals $\frac{k_{r}}{d^{2}}$ for $d < d_{max}$, and 0 otherwise. Here, $k_{r}$ represents a tunable repulsion gain and $d_{max}$ defines the spatial activation margin beyond which the obstacle exerts no influence. 

During the RL training phase, the policy $\pi_{\theta}$ is optimized, while the online surfel reconstruction and the geometric fabric are non-trainable components. At each step, the surfel map is updated from the incoming RGB-D observation, the distance field and gradient are queried at the current sphere centers, the repulsive accelerations are assembled, and the fabric resolves the policy-driven attraction and the sensor-driven repulsion into a safe, configuration-space acceleration $\ddot{q}$. By offloading the complexity of real-time obstacle avoidance entirely to the geometric fabric, the RL agent can focus its capacity exclusively on complex task completion.

\section{Results}
\label{sec:results}

In this section, we validate our claims.
Section~\ref{result_recon} validates the computational efficiency and reconstruction quality of our method, demonstrating its viability for both RL training loops and real-time policy inference. 
Section~\ref{result_manip} evaluates our contributions in the context of collision-rich manipulation tasks. We show that integrating our reconstruction system during both training and deployment allows policies to tackle geometrically complex tasks, enhances robustness against train-test shifts, and allows sim-to-real deployment with improved collision avoidance.

\subsection{Batched 3D Reconstruction}
\label{result_recon}

We evaluate our batched reconstruction method (Sec.~\ref{subsec:recon}) in terms of GPU memory, per-step runtime, 
and reconstruction quality. To support this evaluation, we generate a tabletop manipulation dataset in \textit{Isaac Lab Arena}\footnote{\url{https://github.com/isaac-sim/IsaacLab-Arena}} featuring a robot arm in a cluttered tabletop workspace, with an orbiting camera, a setup typical of manipulation-focused RL training.

\begin{figure*}[t]
  \centering
  \includegraphics[width=\textwidth]{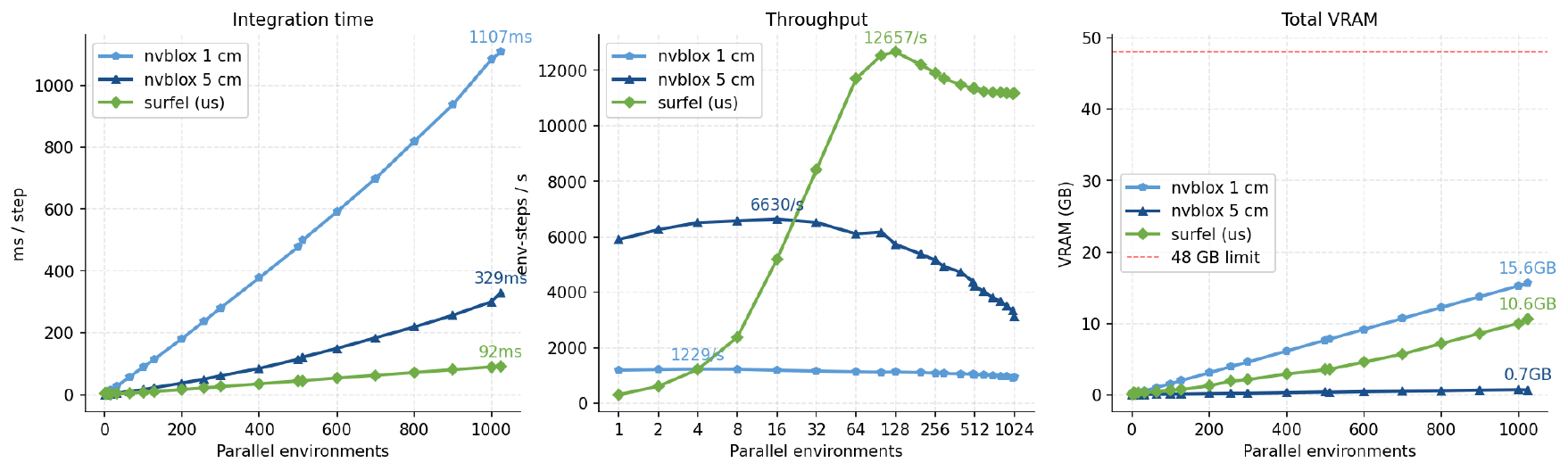}
  \caption{Scaling comparison of \nvblox{} (non-batched; looped over environments) and our approach (batched) as the number of parallel environments increases, showing that batched mapping is substantially more efficient at larger environment counts.}
  \vspace{-10pt}
  \label{fig:scaling_experiment}
\end{figure*}

\paragraph{Memory and Runtime Performance} We evaluate computational efficiency by comparing our method to the widely used, real-time but non-batched \nvblox~\cite{millane2024nvblox}.
We compute reconstructions for increasing numbers of parallel environments.
All experiments run on a single NVIDIA L40 GPU using 500 frames per environment on the tabletop dataset (each environment is a copy of the same sequence), with RGB-D frames downsampled to $80{\times}60$ (matching teacher training; see Sec.~\ref{sec:timings-breakdown}).
To measure parallel reconstruction performance, we run \nvblox{} sequentially across environments, while our method integrates all environments in a single batched GPU workload.
To capture different points of the speed-accuracy tradeoff, we run \nvblox{} at 1 and 5\,cm voxel sizes.
Fig.~\ref{fig:scaling_experiment} reports integration times and GPU memory averaged over 500 frames as the number of parallel environments increases from 1 to 1024. 
Our method surpasses \nvblox{} (5\,cm) in throughput for environment counts above 32 and \nvblox{} (1\,cm) above 8 environments. At 1024 environments, it delivers 3.3$\times$ higher throughput than \nvblox{} (5\,cm) and 12.1$\times$ higher throughput than \nvblox{} (1\,cm), while using 10.6\,GB of GPU memory compared to 15.7\,GB for \nvblox{} (1\,cm) and 0.75\,GB for \nvblox{} (5\,cm).
This scaling advantage arises because batching all environments into a single GPU workload is substantially more efficient than looping, which causes runtime to increase linearly with environment count.
The efficiency of our system at high environment counts enables in-training reconstruction, as we shall~see.

\paragraph{Reconstruction Accuracy} Table~\ref{tab:reconstruction_accuracy} reports reconstruction quality on
the dataset described above, using the same camera trajectory,
ground-truth depth, and identical reconstruction
settings as in the scaling evaluation. We compare against
\nvblox{} and report $accuracy$ and $completion$ at a
$3\,\mathrm{cm}$ threshold, computed against the ground-truth mesh. 
For $accuracy$, we measure the point-to-surface distance from each reconstruction point (surfel center for ours; mesh vertex for \nvblox{}) to the ground-truth mesh. For $completion$, we measure the distance from each ground-truth mesh vertex to its nearest reconstructed surface element (nearest surfel disk for ours; nearest mesh vertex for \nvblox{}).
Compared to \nvblox{} at 5\,cm and 1\,cm voxel size, our surfel reconstruction achieves the highest accuracy ($99.8\%$) with a completion ratio ($85.4\%$) that lies between the two \nvblox{} baselines ($66.9\%$ at 5\,cm; $93.3\%$ at 1\,cm). Overall, our method achieves reconstruction quality comparable to that of \nvblox{}, while achieving the performance required for highly batched RL environments.

\begin{table}[t]
  \centering
  \small
  \caption{Static reconstruction performance on the \textit{Isaac Lab Arena} tabletop dataset. Metrics are reported as fractions at a $3\,\mathrm{cm}$ threshold (higher is better).}
  \label{tab:reconstruction_accuracy}
  \setlength{\tabcolsep}{6pt}
  \begin{tabular}{lcc}
    \toprule
    \textbf{Method} & \textbf{Accuracy} & \textbf{Completion} \\
    \midrule
    \nvblox{} (5 cm) & 66.7 \% & 66.9 \% \\
    \nvblox{} (1 cm) & 99.4 \% & \textbf{93.3} \% \\
    Surfel (Ours) & \textbf{99.8} \% & 85.4 \% \\
    \bottomrule
  \end{tabular}
  \vspace{-5pt}
\end{table}

\subsection{Manipulation Under Geometric Constraints}

In this section, we aim to validate our claims that a) we can train policies that utilize reconstruction during training and transfer sim-to-real, b) this approach produces policies that can be applied to novel, collision-dense tasks where other approaches fail, and c) that these policies are more robust to test time variation.

\label{result_manip}

\subsubsection{Experimental Setup and Training Strategy}

Following~\cite{lum2024dextrah, singh2024dextrah}, we employ a teacher-student distillation framework.
To ensure robust sim-to-real transfer, the teacher training utilizes Automatic Domain Randomization (ADR)~\cite{akkaya2019solving}. ADR progressively expands the randomization bounds for target objects, fixtures (e.g., shelves and boxes), and clutter based on policy success thresholds. Following convergence, the privileged teacher is distilled into a stereo-RGB student policy via DAgger~\cite{ross2011reduction}.
To prevent the robot or the manipulated object from registering as obstacles, we mask RGB-D images before they're fused into the map. In simulation, we use ground-truth segmentations for this masking. In the real world, target objects are masked using Florence-2~\cite{xiao2024florence} combined with SAM2~\cite{ravi2024sam2}, and the robot is masked by projecting its forward-kinematic collision spheres into the images.

\textbf{Robot setup:} The platform consists of a Franka arm with a
Robotiq gripper (similar to DROID~\cite{khazatsky2024droid}; see Fig.~\ref{fig:teaser}). The
setup provides stereo RGB images, which are provided to the deployed
policy, and stereo depth images are used to perform the reconstruction
passed to geometric fabrics.

\begin{figure}[t]
  \centering
  \begin{tabular}{@{}c c@{}}
    \includegraphics[width=0.48\linewidth]{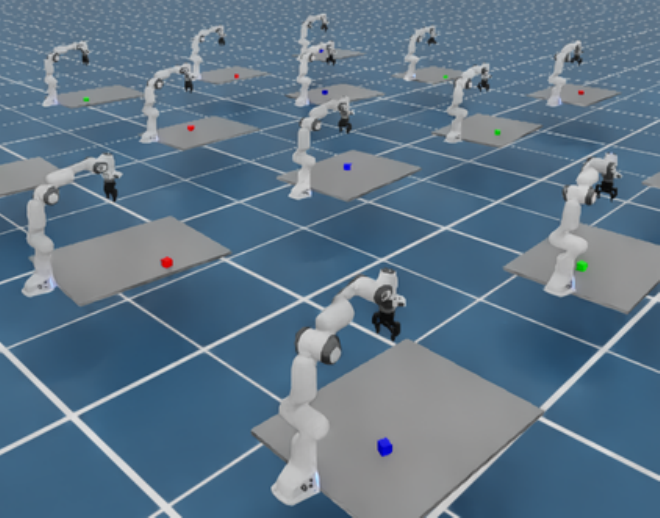} &
    \includegraphics[width=0.48\linewidth]{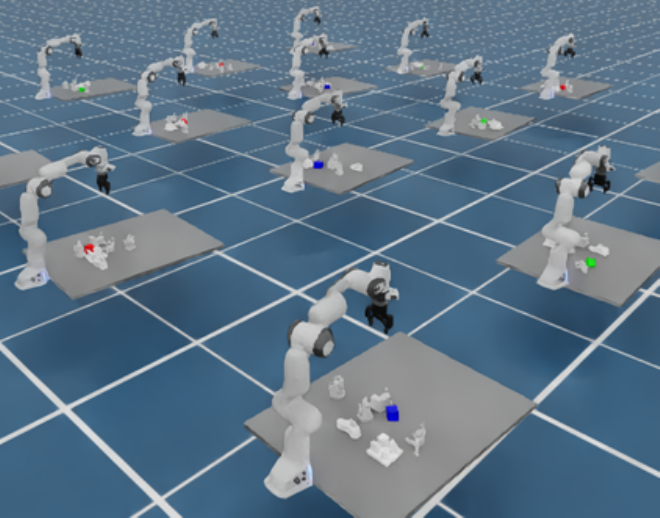} \\
    {Pick from Tabletop} & {Pick from Cluttered Tabletop} \\[0.25em]
    \includegraphics[width=0.48\linewidth]{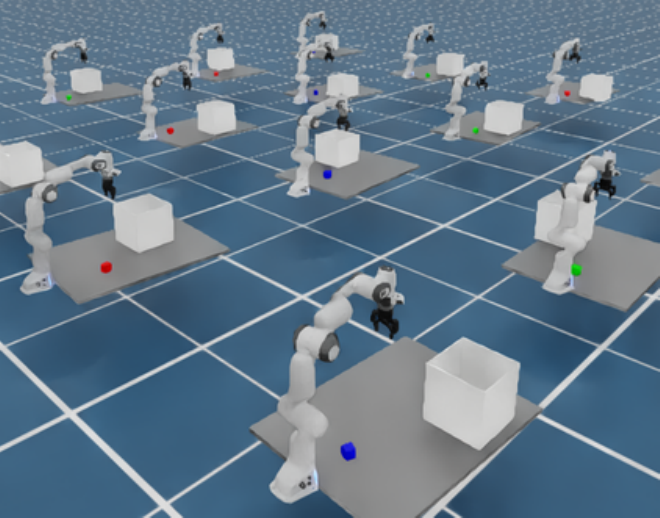} &
    \includegraphics[width=0.48\linewidth]{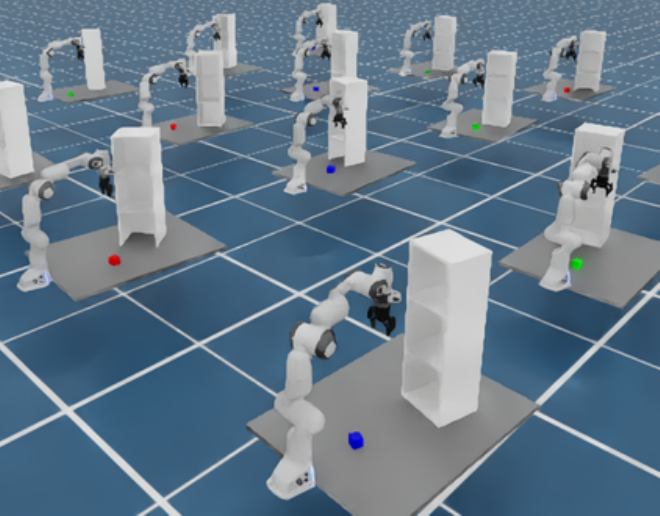} \\
    {Place in Box} & {Place in Shelf}
  \end{tabular}
  \caption{\textbf{Evaluation environments.} Four simulated manipulation scenes used to evaluate policies on progressively geometrically constrained tasks that require obstacle avoidance. We also replicate these scenes/tasks in the real world.}
  \vspace{-5pt}
  \label{fig:evaluation_environments}
\end{figure}

\textbf{Evaluation environments:} We evaluate policies on four environments in simulation (see Fig.~\ref{fig:evaluation_environments}), which we also replicate in the real world. These environments are progressively geometrically complex and require obstacle avoidance for success. We compare the capability and robustness of policies with and without the surfel fabric:

\begin{itemize}[leftmargin=*, noitemsep]
    \item \textbf{Pick from Tabletop:} A baseline task requiring no collision avoidance: pick a target object (a cube, in our experiments) from a table and lift it to a goal pose.
    \item \textbf{Pick from Cluttered Tabletop:} As above, but with additional
    distractor objects cluttering the tabletop. The policy must
    navigate around clutter while grasping the target.
    \item \textbf{Place in Box:} Pick the target object and
    place it into a box. Success requires collision
    avoidance both during grasping and during insertion.
    \item \textbf{Place in Shelf:} Similar to the box
    task, but the target must be placed into an open shelf.
\end{itemize}

Across tasks, the poses of the cube and any present clutter or fixture (box and shelf) are randomized.

\subsubsection{Baseline Methods} 

We compare surfel fabrics against a set of baselines that vary in what scene information is available to the teacher and/or the student.

\textbf{Common setup:} All methods use the same task definitions and rewards. During teacher training, the teacher policy receives proprioception and ground-truth target object and goal poses (no vision observations), and we distill these ground-truth pose signals to the student via auxiliary losses following~\cite{singh2024dextrah}. The deployed student receives proprioception and stereo RGB only.

\begin{itemize}[leftmargin=*, noitemsep]
    \item \textbf{(i) Primitive fabric:} No additional scene geometry beyond stereo RGB.
    \item \textbf{(ii) Primitive fabric with ground-truth obstacle/fixture poses:} Teacher additionally receives ground-truth poses of clutter objects and fixtures (shelf/box), distilled to the student via auxiliary losses (no ground-truth pose available to the student at test time).
    \item \textbf{(iii) Policy with pointcloud observation and primitive fabric:} Teacher and student policies additionally observe the scene as a pointcloud: depth images are unprojected, cropped to the workspace, and subsampled to 512 points, which a small PointNet-style encoder~\cite{qi2017pointnet} embeds into a 64-dimensional policy input. Scene geometry influences behavior only through the learned policy, not through the fabric.
    \item \textbf{(iv) Surfel fabric (ours):} Collision avoidance is handled by the scene-aware fabric interface, queried from our surfel reconstruction (the policy receives no explicit obstacle observations).
\end{itemize}

In the following subsections, we evaluate teacher-training success, evaluate robustness to unseen obstacles, report real-robot deployment, and analyze training times.

\subsubsection{Teacher Training Evaluation}

We evaluate teacher training for each method's ability to produce policies that complete the tasks. We use ADR to progressively increase scene difficulty via fixture/obstacle randomization: $\mathrm{ADR}=0$ is the easiest setting (no randomization), and $\mathrm{ADR}=50$ is the maximum randomization. We treat making progress beyond $\mathrm{ADR}>0$ as a proxy for the teacher starting to learn the task, and reaching $\mathrm{ADR}=50$ as a proxy for finishing learning with a policy that succeeds under full randomization. Both measurements therefore indicate learning success on the task. For each method--task pair, we run 5 different teacher training seeds and report how many seeds reach $\mathrm{ADR}>0$ and $\mathrm{ADR}=50$ in Table~\ref{tab:adr_results}.

Table~\ref{tab:adr_results} shows that our proposed method is the only method in which the teacher policy is able to successfully learn the behaviors to complete all 4 tasks.
In particular, for the last two tasks, \textit{Place-in-Box} and \textit{Place-in-Shelf}, no other methods complete teacher training.
Primitive fabrics do not complete training on these tasks that include complex collision geometry: without ground-truth supervision or vision observation, the teacher has no scene information (no clutter or fixture state).
We also find that adding ground-truth scene information does not help. This aligns with intuition; the policy only receives obstacle/fixture poses but not geometry, so it still lacks the geometric context needed for reliable collision avoidance as randomization increases. As a result, primitive fabrics fail to reach $\mathrm{ADR}=50$.
Pointcloud-based policies, which observe the scene geometry, also fail at the geometrically complex tasks.
Success in this case would require the policy to learn to interpret a pointcloud to enable obstacle avoidance under sparse rewards.
In contrast, surfel fabrics do not have this requirement; the fabric provides collision avoidance, and the policy only needs to learn the goal-directed behavior.

We also observe that surfel fabrics succeed in fewer seeds on the tasks requiring no (\textit{Pick-from-Tabletop}) or only limited (\textit{Pick-from-Cluttered-Tabletop}) obstacle avoidance; this is a matter for future investigation, and we hypothesize that our current fabric tuning better supports avoidance of larger obstacles (e.g., shelves and boxes) than navigation around small, densely packed clutter.

\begin{table*}[t]
  \centering
  \caption{Teacher training success across 5 seeds. Each cell reports the number of seeds reaching $\mathrm{ADR}>0$ / $\mathrm{ADR}=50$ as ``$n/m$'' (out of 5). Our method is the only one to reach $\mathrm{ADR}=50$ on all tasks (i.e. to successfully learn the behaviors).}
  \vspace{3pt}
  \label{tab:adr_results}
  \small
  \setlength{\tabcolsep}{5pt}
  \begin{tabular}{lcccc}
    \toprule
    \textbf{Task} & \textbf{Primitive (no-GT)} & \textbf{Primitive (GT)} & \textbf{Pointcloud} & \textbf{Surfel (Ours)} \\
    \midrule
    Pick from Tabletop              & 2/2 & 2/2 & \textbf{3}/\textbf{2} & 2/1 \\
    Pick from Cluttered Tabletop & 2/2 & 2/2 & \textbf{4}/\textbf{3} & 1/1 \\
    Place in Box      & 0/0 & 0/0 & 0/0 & \textbf{3}/\textbf{2} \\
    Place in Shelf    & 1/0 & 1/0 & 0/0 & \textbf{2}/\textbf{2} \\
    \bottomrule
  \end{tabular}
  \vspace{-8pt}
\end{table*}

\subsubsection{Robustness Evaluation}

In this section we evaluate our claim that surfel fabric-based policies are more robust to changing scene geometry, by introducing an obstacle that is absent during training.

\begin{figure*}[t]
    \centering
    \includegraphics[width=\textwidth]{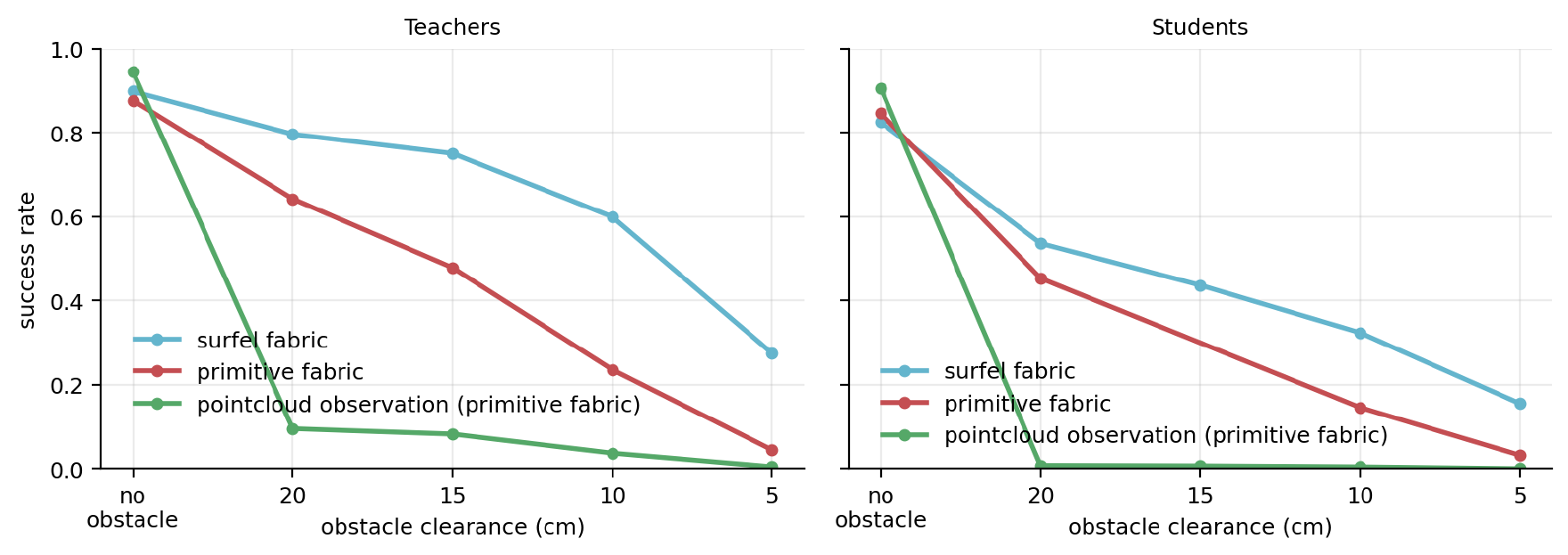}
    \caption{Robustness to an unseen obstacle at test time. A cylinder is placed in the workspace at varying clearances from the target object. The policies were trained on the \textit{Pick-from-Tabletop} task with no such obstacle. We plot collision-free success as a function of clearance. Surfel fabrics are consistently more robust than primitive fabrics.}
    \label{fig:unseen_obstacle}
    \vspace{-5pt}
\end{figure*}

Starting from policies trained on \textit{Pick-from-Tabletop}, we introduce a cylindrical obstacle at test time and vary its radius (5--20\,cm), height (10--50\,cm), and clearance to the target cube (5--20\,cm) (see Fig.~\ref{fig:unseen_obstacle}). Averaged over obstacle randomizations, the surfel fabric achieves 61\% collision-free success, compared to 35\% for primitive fabrics and 6\% for a pointcloud-observation policy, a 1.7$\times$ improvement over primitives and an 10$\times$ improvement over pointcloud observations. The advantage also survives distillation. The surfel student reaches 36\% collision-free success, compared to 21\% for the primitive student (1.7$\times$), and below 1\% for the pointcloud student.

Overall, these results suggest that policies based on surfel fabrics are more robust to changes in scene geometry at test time. This aligns with intuition: collision avoidance is induced by an explicit, test time scene representation, rather than relying on obstacle-specific avoidance being learned during training from sparse rewards.

\subsubsection{Real-Robot Deployment Evaluation}

\begin{figure}[t]
  \centering
  \includegraphics[width=0.24\linewidth]{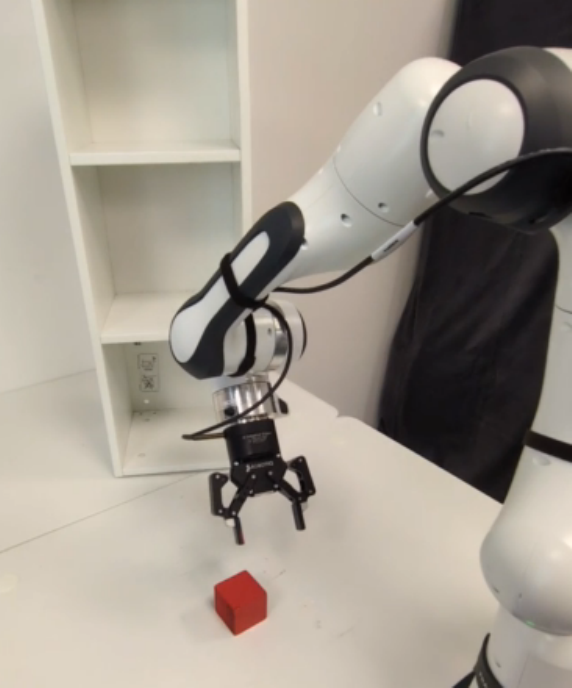}\hfill
  \includegraphics[width=0.24\linewidth]{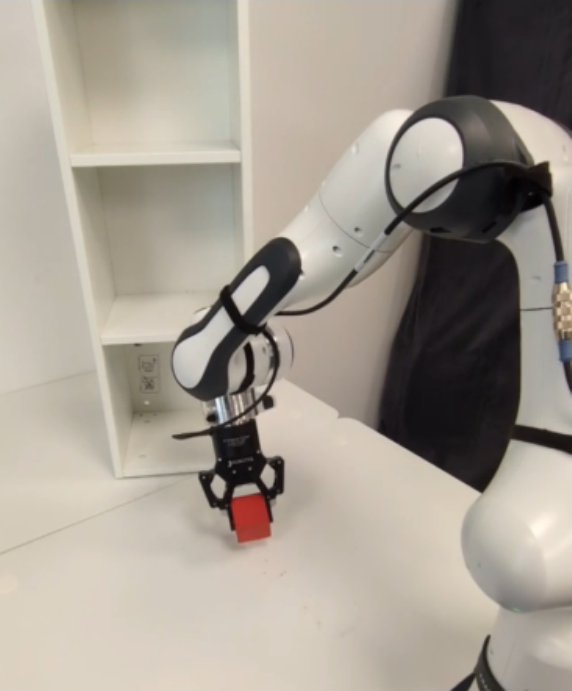}\hfill
  \includegraphics[width=0.24\linewidth]{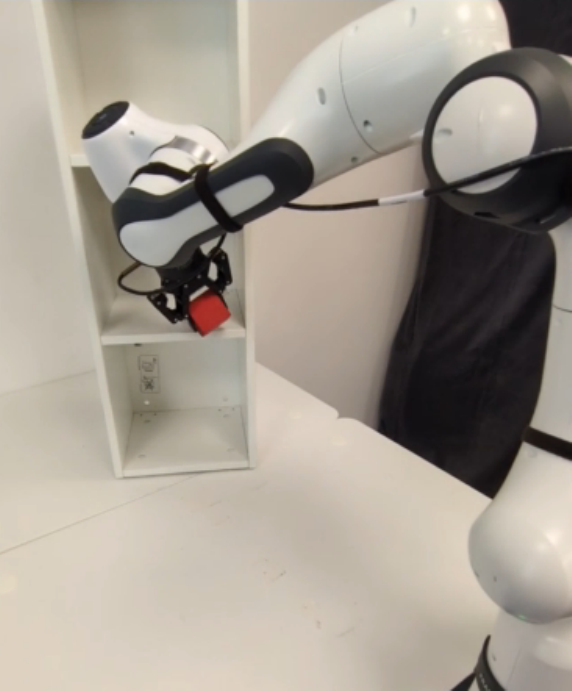}\hfill
  \includegraphics[width=0.24\linewidth]{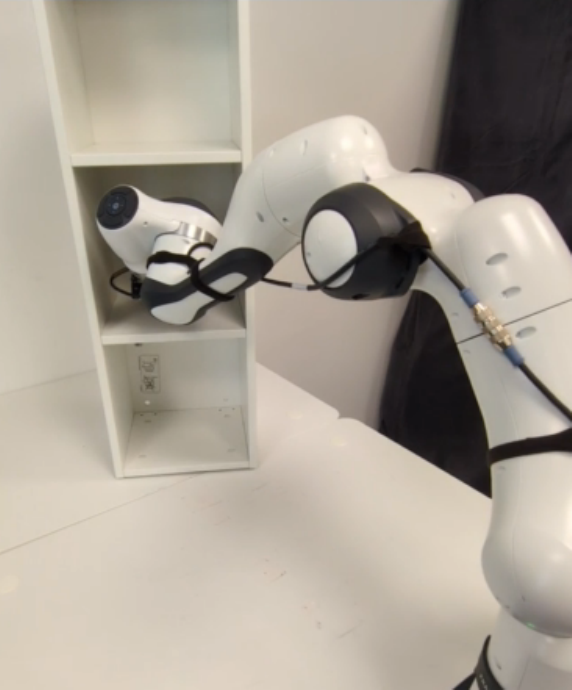}
  \vspace{3pt}
  \caption{\textbf{Place in Shelf Deployment.} Frames showing a successful placement of a cube into the shelf on our real-robot setup (\emph{left-to-right}). Repulsion from the scene-aware fabric allows the policy to direct the robot into the confined shelf compartment without collision.}
  \label{fig:shelf_deploy_sequence}
\end{figure}

In this section, we aim to validate our claim that our method allows extension of geometry fabric-based policies to real-world tasks involving complex scene geometry.
We deploy the surfel fabric student policy on our DROID-based robot across all four tasks and observe successful deployments in all cases. Table~\ref{tab:real_robot_success} reports the real-robot success rate over 10 trials per task (see Fig.~\ref{fig:shelf_deploy_sequence} for an image sequence from a rollout of the \mbox{\textit{Place-in-Shelf}} task).
A main failure mode is cube pickup failure (and, in \mbox{\textit{Place-in-Box}}, early cube release), which we believe is addressable via reward tuning and not primarily due to surfel fabrics.
These real-world experiments show that our system is effective at enabling sim-to-real deployment for tasks in which the baseline methods fail.

\begin{table}[t]
  \centering
  \small
  \caption{Real-robot success rates for surfel fabric policies (10 trials per task).}
  \vspace{3pt}
  \label{tab:real_robot_success}
  \setlength{\tabcolsep}{6pt}
  \begin{tabular}{lc}
    \toprule
    \textbf{Task} & \textbf{Success rate} \\
    \midrule
    Pick from Tabletop              & 70 \% \\
    Pick from Cluttered Tabletop & 60 \% \\
    Place in Box      & 30 \% \\
    Place in Shelf    & 100 \% \\
    \bottomrule
  \end{tabular}
  \vspace{3pt}
\end{table}

\vspace{10pt}

\subsubsection{Qualitative Real-Robot Case Studies}
\label{sec:qualitative_real_robot}

Beyond the success rates reported in Table~\ref{tab:real_robot_success}, we present two qualitative real-robot experiments.

\begin{figure}[t]
  \centering
  \includegraphics[width=0.32\linewidth]{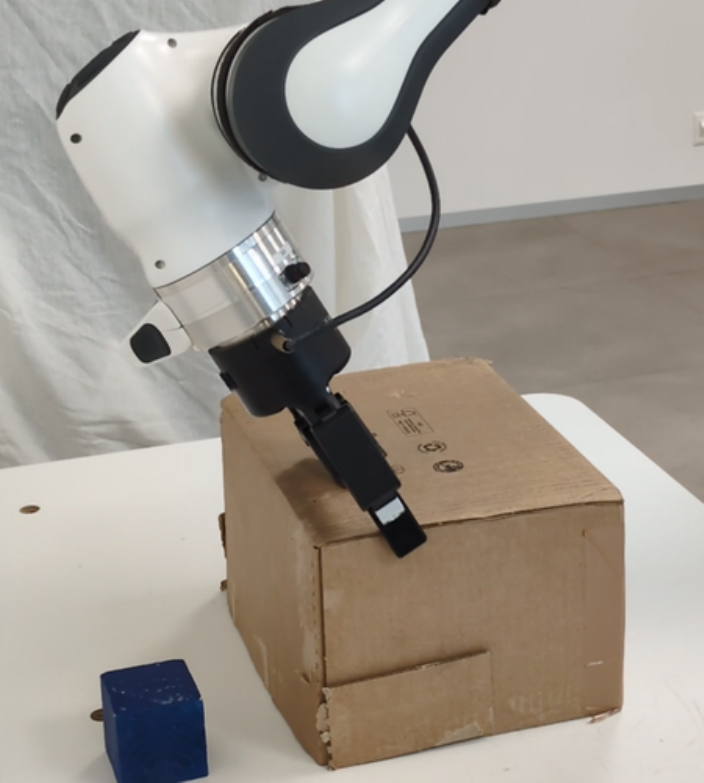}\hfill
  \includegraphics[width=0.32\linewidth]{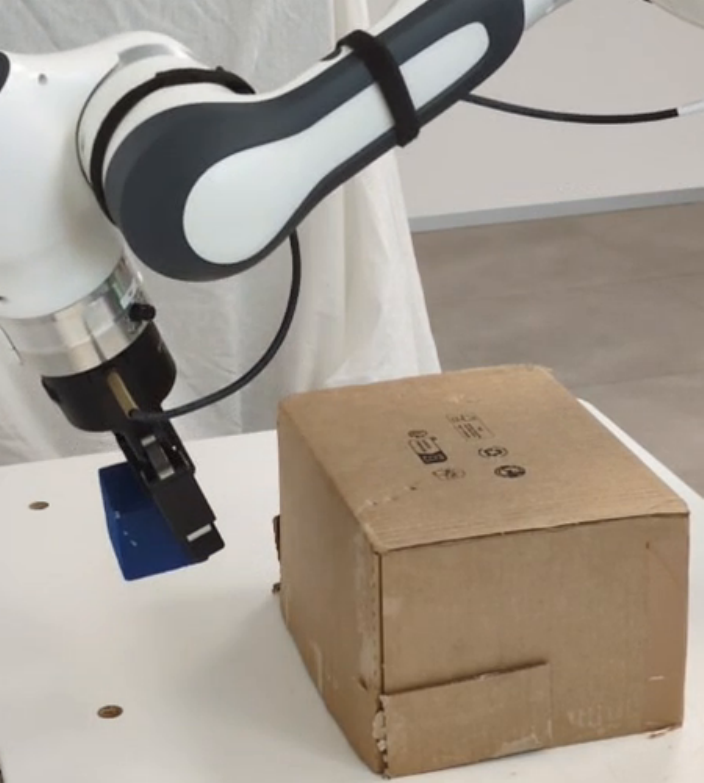}\hfill
  \includegraphics[width=0.32\linewidth]{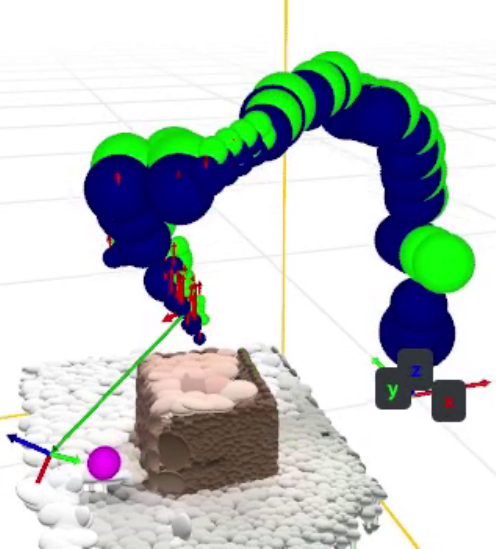}
  \vspace{3pt}
  \caption{\textbf{Unseen obstacle at deployment.} A cardboard box, absent from training, blocks the path to the target cube. Without surfel fabrics, the gripper collides with the box (\emph{left}). With surfel fabrics, the task succeeds (\emph{middle}); the end-effector goal predicted by the policy (shown as coordinate frame) is converted into a collision-avoidant trajectory by repulsive forces (red arrows) from the reconstruction (\emph{right}).}
  \vspace{-5pt}
  \label{fig:real_unseen_obstacle}
\end{figure}

\noindent\textbf{Generalization to unseen obstacles on real-robot.}
We deploy a \textit{Pick-from-Tabletop} student policy with surfel fabrics and introduce a physical obstacle absent from training, mirroring the simulated unseen obstacle experiment of Fig.~\ref{fig:unseen_obstacle}. The policy completes the pickup task while the fabric steers the arm around the obstacle, with no retraining or obstacle specific reward shaping (see Fig.~\ref{fig:real_unseen_obstacle}).

\begin{figure}[t]
  \centering
  \includegraphics[height=2.4cm,keepaspectratio]{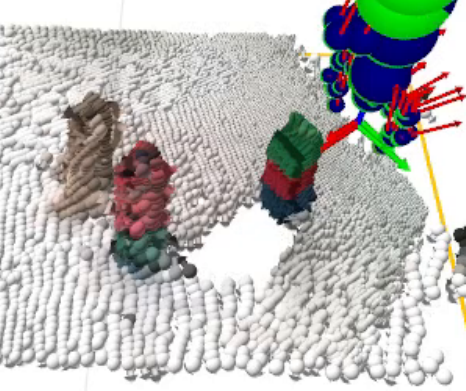}\hfill
  \includegraphics[height=2.4cm,keepaspectratio]{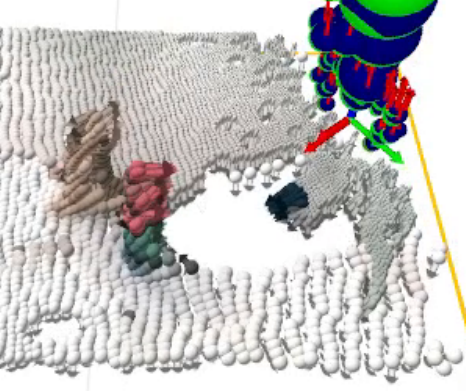}\hfill
  \includegraphics[height=2.4cm,keepaspectratio]{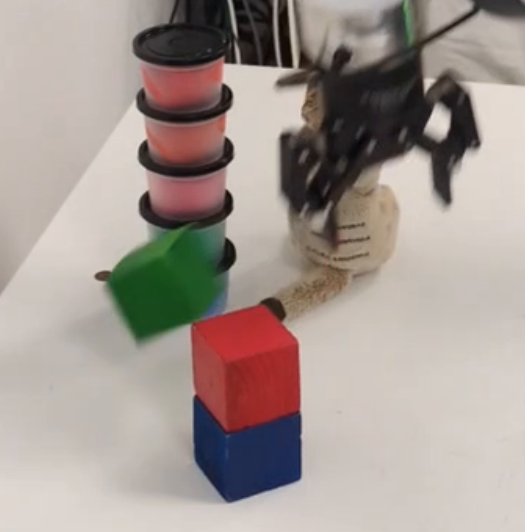}
  \caption{\textbf{Collision avoidance under occlusion.} Our accumulated reconstruction retains an obstacle (a stack of cubes) even while it is temporarily occluded by the robot gripper (\emph{left}), generating repulsive forces (red arrows) that steer the arm around it during a point-to-point command through the obstacles. With instantaneous per-frame reconstruction, the obstacle is forgotten once out of view (\emph{middle}), removing repulsive geometry and causing collision (\emph{right}).}
\label{fig:instantaneous_vs_accumulated}
\end{figure}

\noindent\textbf{Collision avoidance with accumulated vs. per-frame reconstruction.}
To test surfel fabric's repulsion mechanism in isolation, we remove the learned policy and command a fixed point-to-point motion. We test two reconstruction modes. Firstly, a per-frame reconstruction, which has no memory of previously seen geometry, and secondly, our method, which accumulates a reconstruction over time. Our experiments show that with our method the fabric retains currently-unseen geometry and repels the arm around it, while with an instantaneous view the arm can collide with objects that have left the camera's field of view (see Fig.~\ref{fig:instantaneous_vs_accumulated}).

\subsubsection{Training Timings Breakdown}\label{sec:timings-breakdown}

To keep teacher training fast, we use a low input resolution (80$\times$60) for surfel mapping and use two cameras to maximize workspace coverage. Teacher training takes 385\,ms per environment step (measured on a single GPU with 1024 parallel environments), and is dominated by three similarly costly components: rendering (98\,ms), mapping (93\,ms), and image masking (90\,ms). The reported times are averaged over 1\,hour of training. Total GPU memory usage is 23.5\,GB, split across the primitive baseline fabric (9.5\,GB), rendering (7.6\,GB), and surfel fabrics (6.5\,GB). Total host RAM usage is 88\,GB, and is largely driven by rendering (75.6\,GB).

\vspace{5pt}

Across all runs, we use 8192 environments in total. On \textit{Pick-from-Tabletop} and \textit{Pick-from-Cluttered-Tabletop}, primitive fabrics reach full task randomization at $\mathrm{ADR}=50$ in 16--24\,h (6.5--6.6k rollouts), the pointcloud-observation baseline in 39--50\,h (7.7--7.8k), and surfel fabrics in 32--38\,h (7.6--8.0k). On \textit{Place-in-Box} and \textit{Place-in-Shelf}, only surfel fabrics successfully learn, reaching $\mathrm{ADR}=50$ in 37--39\,h (7.9--8.1k rollouts). Due to increased memory consumption from rendering and perception, surfel fabrics and the pointcloud policy use 8 NVIDIA L40 GPUs (1024 environments per GPU), whereas primitive fabrics run on a single GPU. Overall, surfel fabrics train in a comparable number of steps to primitive fabrics, but require more wall-clock time due to rendering and mapping overhead. We expect this gap to narrow with further optimizations to rendering and mapping.

\section{Future work, Limitations and Conclusion}

\label{sec:conclusion}

Our method integrates real-time reconstruction into massively parallel reinforcement learning to enable learning on top of scene-aware controllers. We extend geometric fabrics by replacing static, hand-crafted geometric primitives with a real-time, sensor-derived map, and enable collision avoidance without compromising the high computational throughput required for modern RL training. Empirical evaluations across complex, collision-dense manipulation tasks demonstrate that policies trained with our method exhibit superior success in completing complex tasks compared to primitive-based baselines, are more robust to the introduction of novel geometry at test time, and show successful sim-to-real transfer. 
A current limitation is that we only evaluate static cameras where persistent occlusions can remain unmapped. Looking ahead, we aim to study dynamic cameras (e.g., head and wrist-mounted). 
We also plan to leverage the fully differentiable nature of the tensorized method for end-to-end gradient flow from the representation to upstream perception modules (e.g., image encoders).
By eliminating reliance on manual world modeling in favor of dynamic spatial representations, our contributions provide a scalable foundation for deploying collision-aware robotic control in unstructured, real-world environments.

\vspace{-2pt}




\bibliographystyle{icra_template/bib/IEEEtran}
\bibliography{icra_template/bib/IEEEabrv,references}

\clearpage
\appendix
\section{Supplemental Material}
\label{sec:appendix}
This appendix provides supplemental material that supports the results in
Sec.~\ref{sec:results}, including extended versions of the qualitative figures and an additional
benchmarking breakdown plot. Sec.~\ref{app:dataset} describes the tabletop
reconstruction dataset we generated for benchmarking (see results in
Sec.~\ref{result_recon}) and which we release publicly.

\subsection{Tabletop Reconstruction Dataset}
\label{app:dataset}
To support our evaluation, we constructed a tabletop manipulation
reconstruction dataset in simulation using \textit{Isaac Lab Arena}\footnote{\url{https://github.com/isaac-sim/IsaacLab-Arena}} (see Fig.~\ref{fig:qual_recon} for example reconstructions).
The scenes feature a robot arm in a cluttered tabletop workspace with
an orbiting camera, a setup typical of manipulation-focused RL
training. \textit{Isaac Lab Arena}'s composable environment layer lets us swap in
different clutter object sets while keeping the tabletop and robot
setup fixed, avoiding the need to hand-author each object
combination as a separate environment. The dataset comprises ten 500-frame sequences
($5000$ RGB-D frames in total, $1200{\times}680$ resolution). Each sequence
places a different cluttered object set (e.g., jars, bowls, mugs, or
food items) on the table. We record the ground-truth camera
trajectory and provide a ground-truth mesh trimmed to the
camera-observed surfaces.


\begingroup
\setlength{\stripsep}{40pt}
\vspace{6pt}

\begin{strip}
  \centering
  \begin{tabular}{c@{\hspace{5em}}c}
    \includegraphics[width=0.35\textwidth]{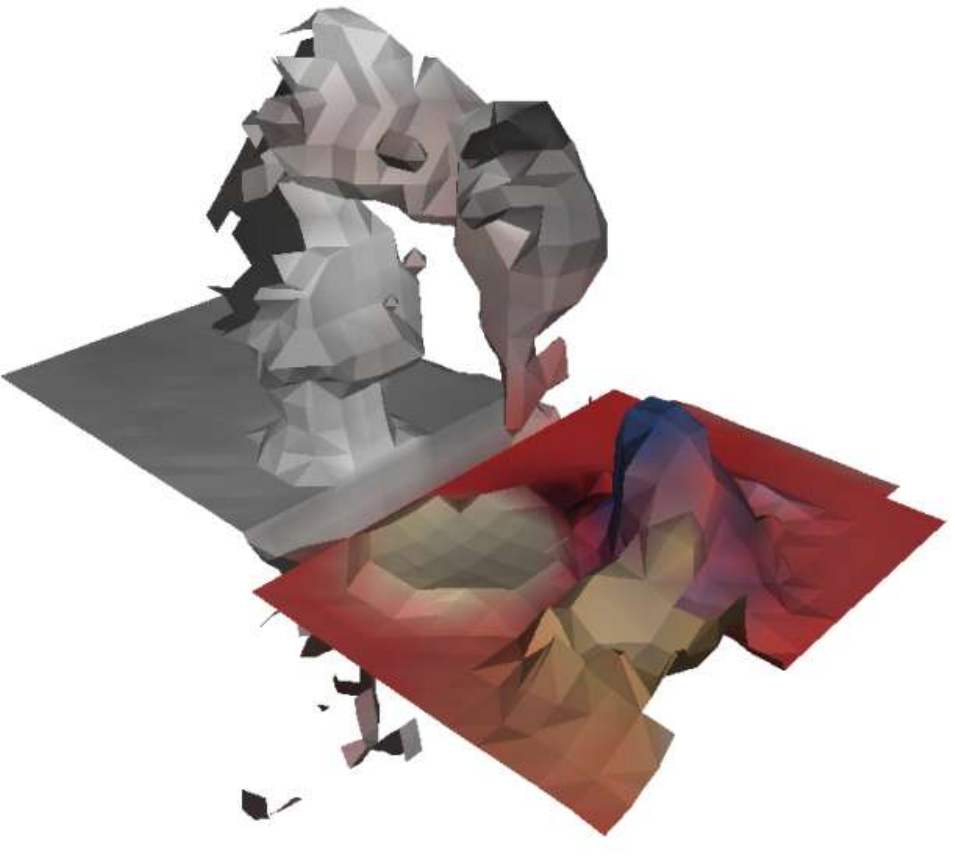} &
    \includegraphics[width=0.35\textwidth]{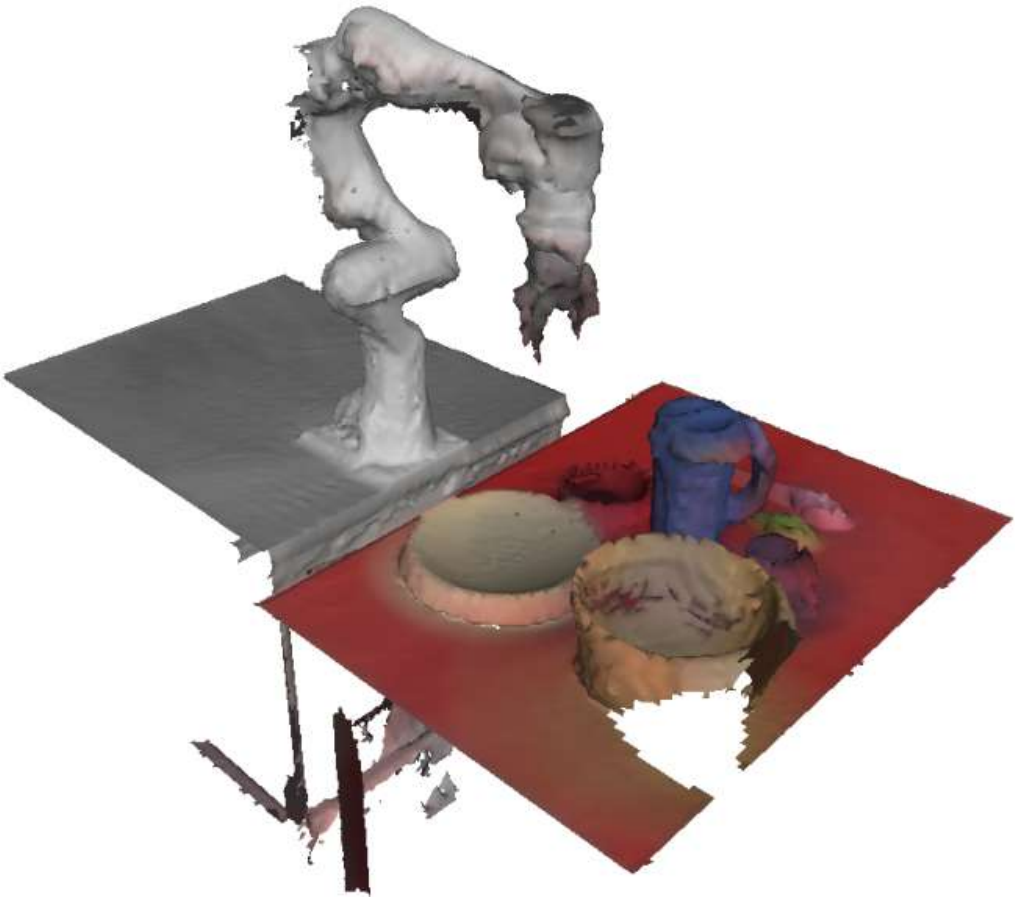} \\
    {\nvblox{} (5\,cm voxels)} &
    {\nvblox{} (1\,cm voxels)} \\[2em]
    \includegraphics[width=0.35\textwidth]{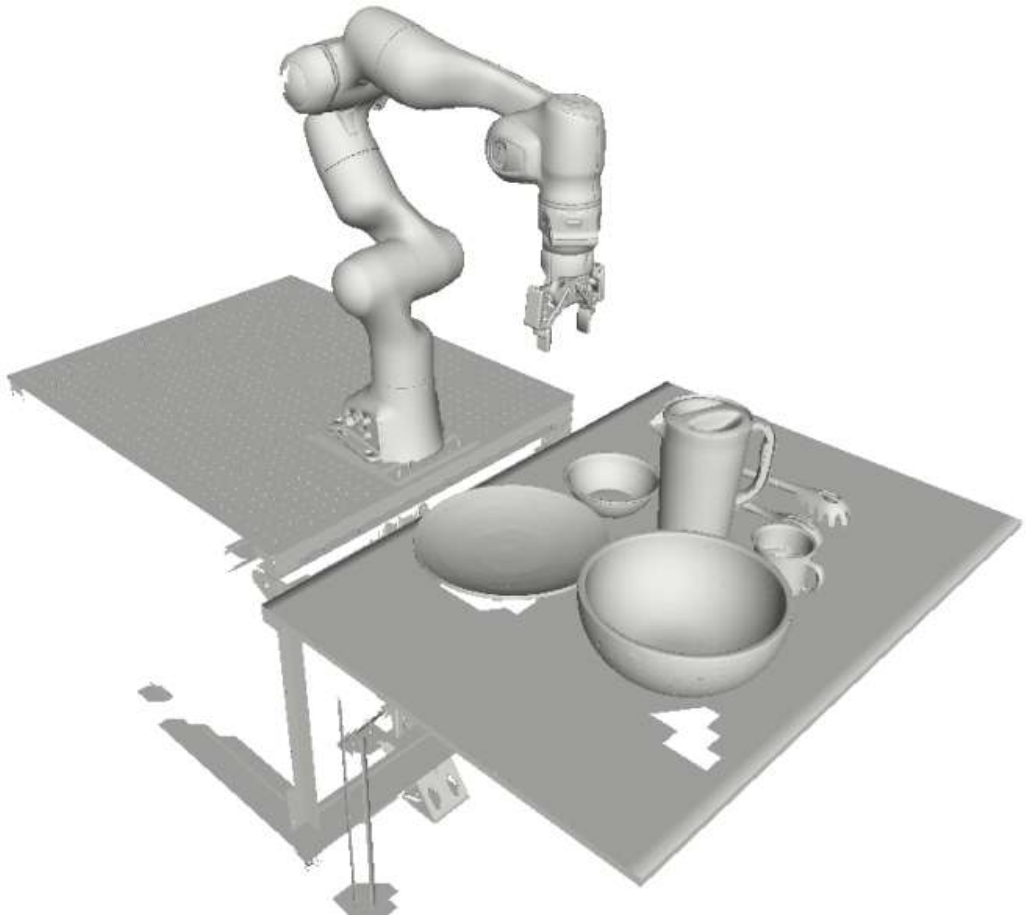} &
    \includegraphics[width=0.35\textwidth]{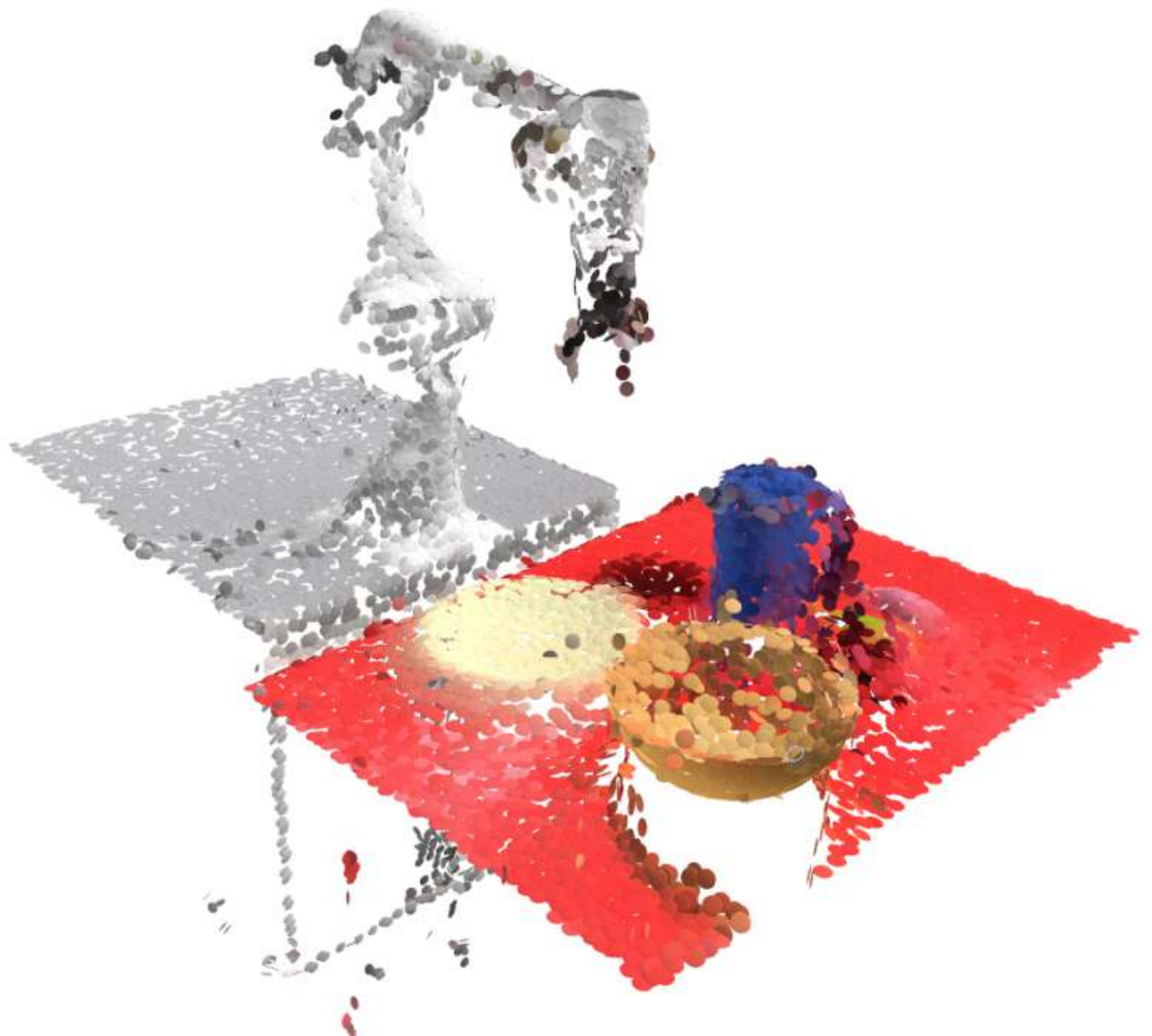} \\
    {GT mesh} &
    {Surfel (Ours)}
  \end{tabular}
  \vspace{2em}
  \captionof{figure}{Comparison of various reconstruction methods a representative
  scene. Top row (left to right): \nvblox{} at 5\,cm voxel size and
  \nvblox{} at 1\,cm voxel size. Bottom row:
  ground-truth mesh and our surfel reconstruction.}
  \label{fig:qual_recon}
\end{strip}
\endgroup




\clearpage

\begingroup
\setlength{\abovecaptionskip}{2pt}
\setlength{\belowcaptionskip}{2pt}

\begin{figure*}[t]
  \centering

  \includegraphics[width=\textwidth,height=0.4\textheight,keepaspectratio]{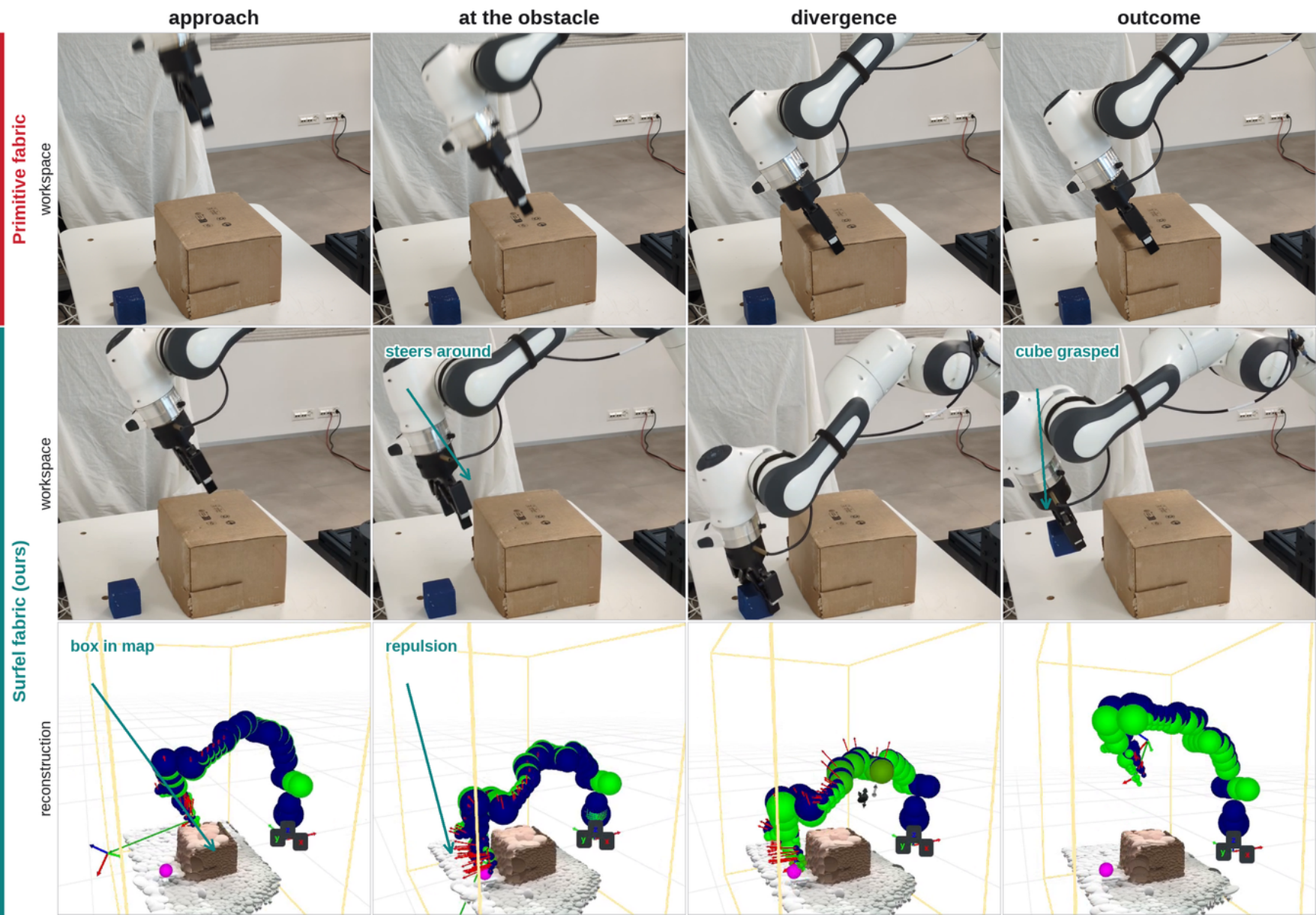}
  \vspace{1pt}
  \captionof{figure}{\textbf{Unseen obstacle at deployment \textit{(extended version of Fig.~\ref{fig:real_unseen_obstacle})}.} A cardboard box absent from training blocks the path to the target cube for pickup. Without surfel fabrics (\emph{top}) the gripper collides with the box. With surfel fabrics (\emph{middle}) the policy is repelled and steers around the box and completes the cube-picking task. The reconstruction used by geometric fabrics to enable this behavior is shown at the bottom. No retraining or obstacle-specific reward shaping was used.}
  \label{fig:real_unseen_obstacle_extended}

  \vspace{6pt}

  \includegraphics[width=\textwidth,height=0.4\textheight,keepaspectratio]{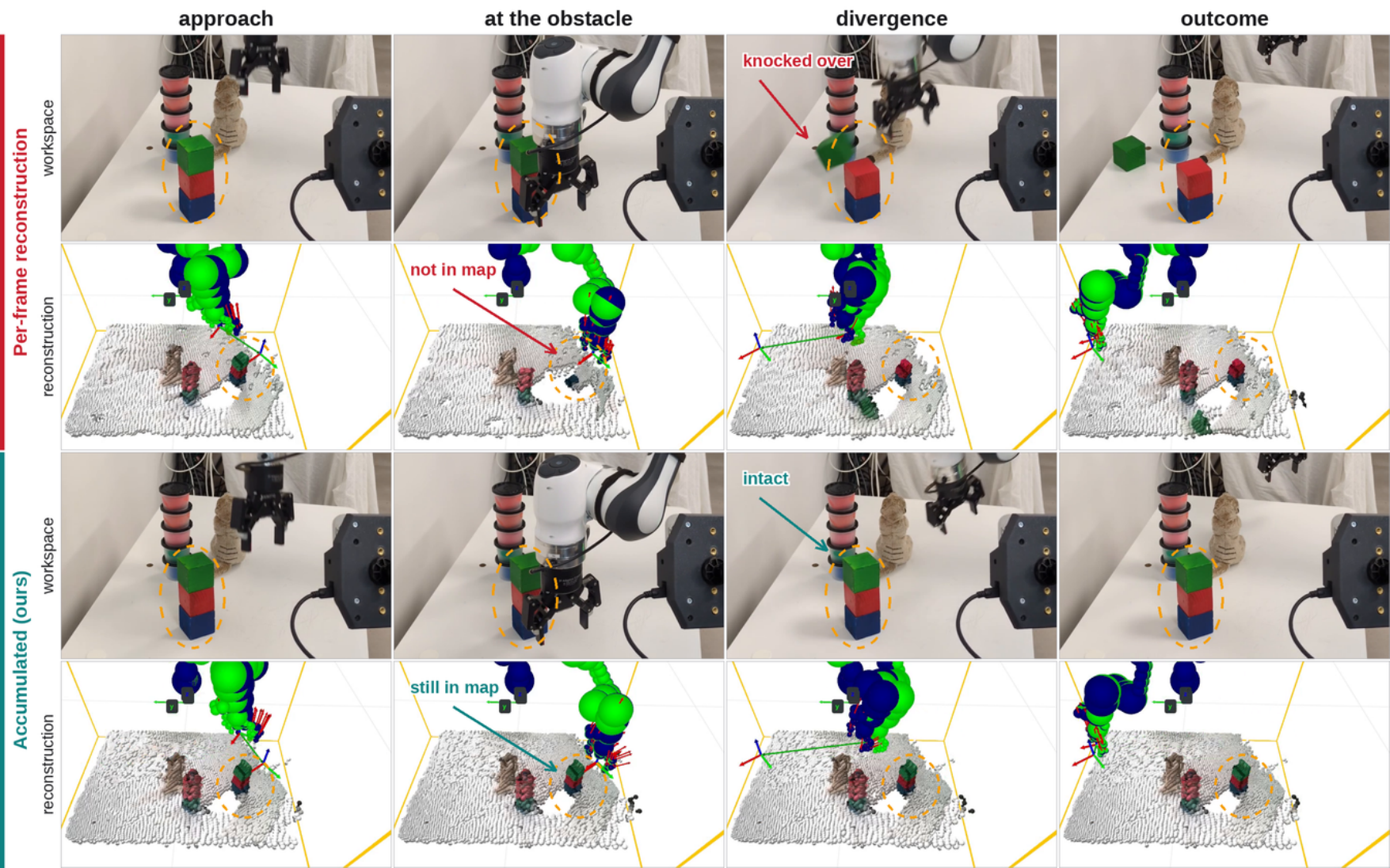}
  \vspace{1pt}
  \captionof{figure}{\textbf{Motion under occlusion \textit{(extended version of Fig.~\ref{fig:instantaneous_vs_accumulated})}.} Identical commanded point-to-point motion under two reconstruction modes (two cameras view the workspace from opposite sides). Rows alternate between workspace view and the surfel map queried by the fabric; the dashed ellipse marks a block tower that must be avoided. \textbf{\emph{Top:}} with per-frame reconstruction the tower is mapped only while directly visible and vanishes as soon as the arm occludes it, leaving no geometry to repel from, and the arm collides with the stack. \textbf{\emph{Bottom:}} accumulated reconstruction retains the tower even while occluded (red arrows show the resulting repulsion), steering the arm around it and leaving the stack intact.}
  \label{fig:instantaneous_vs_accumulated_extended}
\end{figure*}

\endgroup

\newpage

\makeatletter
\setlength{\@dblfptop}{0pt}
\makeatother

\begin{figure*}[!t]
    \centering
    \includegraphics[width=0.9\textwidth]{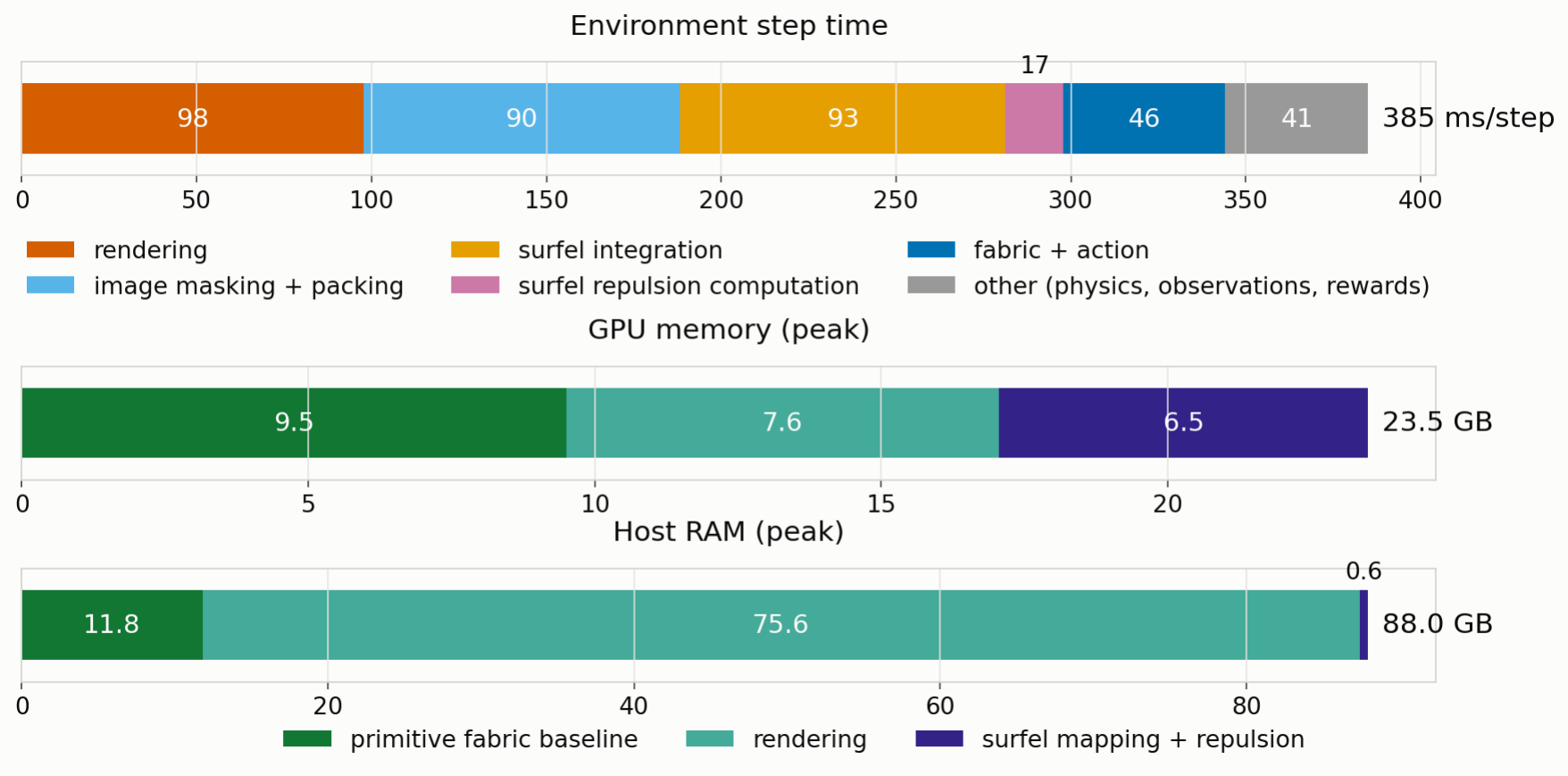}
    \vspace{5pt}
    \caption{\textbf{Timing and memory breakdown \textit{(extending results in Sec.~\ref{sec:timings-breakdown})}.} Step time of teacher training is dominated by rendering, surfel mapping, and image masking. GPU memory is split across the primitive baseline fabric, rendering, and surfel fabrics, while host RAM is largely driven by rendering.}
    \label{fig:timings-breakdown}
\end{figure*}



\end{document}